\documentclass[graybox]{svmult}

\usepackage{mathptmx}       %
\usepackage{helvet}         %
\usepackage{courier}        %
\usepackage{type1cm}        %
\usepackage{makeidx}         %
\usepackage{graphicx}        %
\usepackage{multicol}        %
\usepackage[bottom]{footmisc}%

\usepackage[square,sort,comma,numbers]{natbib}

\makeindex             %

\begin{document}

\title*{Identifying potentiating events in evolutionary search using replay experiments}

\author{Austin J. Ferguson and Alexander Lalejini}

\institute{Austin J. Ferguson \at Grand Valley State University, Allendale, MI, USA; \email{ferguaus@gvsu.edu}
\and Alexander Lalejini \at Grand Valley State University, Allendale, MI, USA; \email{lalejina@gvsu.edu}}
\maketitle

\abstract{
In this work, we introduce analytical replay experiments to the evolutionary computing community. 
Replay experiments originated in the context of laboratory experimental evolution as an empirical approach to identifying potentiating events that increased the likelihood of an observed evolutionary outcome. 
By restarting a population's evolution from different historical time points, replay experiments sample the distribution of what could have evolved from different points in time, which allows us to quantify how a population's potential for different evolutionary outcomes changed as a result of that population's history.  
In this work, we give a step-by-step guide to designing replay experiments for evolutionary computing systems.
We then provide a demonstrative example replay experiment that measures how potentiation for problem-solving success changed in an evolved genetic programming population, showing that increases in potential for success do not necessarily correspond with increases in a population's fitness. 
Broadly, we argue that analytical replay experiments can be a powerful tool for expanding the theoretical foundations of evolutionary computing, and we offer suggestions for promising future research directions enabled by replay experiments.
}

\section{Introduction}
\label{sec:introduction}

Evolutionary computing (EC) harnesses the principles of evolution as a general-purpose search algorithm for solutions to computational problems. 
EC has demonstrated success across a broad range of domains, such as robotics~\citep{doncieux_evolutionary_2015}, repairing bugs in software~\citep{petke_genetic_2018}, and multi-objective optimization~\citep{ehrgott_fifty_2026}.
However, precisely identifying \textit{why} a particular evolutionary search succeeded or failed remains challenging because of the stochastic nature of evolution and the complexity of sophisticated evolutionary computing systems. 
With post-hoc analyses, we can often identify \textit{actualizing} events (e.g., events that conferred a new trait or fitness level) in a population's history.
However, actualizing events alone do not reveal the complete set of critical events that led to a particular end point. 
To use an analogy, we can identify when someone began a new job, but can we measure how past events influenced the likelihood of them being offered and accepting that job? 
In this chapter, we overview how analytical replay experiments can be used to empirically identify \textit{potentiating} events (i.e., events that influenced the likelihood of an actualizing event) in an evolutionary search. 

Analytical replay experiments originated in the context of laboratory experimental evolution studies~\citep{blountHistoricalContingencyEvolution2008, blountContingencyDeterminismEvolution2018}, wherein researchers typically evolve two sets of populations: one set of populations subjected to an experimental treatment (e.g., higher temperatures, resource limitations, \textit{etc.}) and one set of control populations. 
Replay experiments use this same overall experimental design, but instead of manipulating the environment, they vary the amount of shared evolutionary history between population sets. 
Replay experiments start with the historical record of a previously evolved population (the ``historical population'').
For example, in laboratory studies, the historical record might comprise a series of frozen samples from an evolving population of \textit{E. coli}, wherein each sample is taken at a different time during the experiment. 
Using the historical record, we then found a set of evolution replicates from a sample of the historical population taken at generation $N$ and another set of replicates from a sample of the historical population taken at generation $N + K$.
We can then ``replay'' evolution from those two historical starting points many times, and differences in evolutionary outcomes can be attributed to the $K$ generations of additional evolutionary history present in the $N + K$ sample. 
This experimental design allows us to ``replay the tape of life'' as described by Stephen Jay Gould~\citep{gouldWonderfulLifeBurgess1990}. 
That is, we can use replay experiments to sample the distribution of \textit{what could have evolved} from a given point in time.

Analytical replay experiments first gained traction in laboratory experimental evolution studies of microbes, as microbial samples can be frozen for long-term storage and then later revived \citep{lenskiDynamicsAdaptationDiversification1994}.
The seminal use of replay experiments investigated the evolutionary potentiation of citrate metabolism in \textit{E. coli}.
Using replay experiments, Blount \textit{et al.}~\citep{blountHistoricalContingencyEvolution2008} showed that citrate metabolism became increasingly likely to evolve over time as a result of a series of potentiating mutations.  %
Since that initial work, replay experiments have been used to investigate evolvability~\citep{woodsSecondorderSelectionEvolvability2011} and clade extinction in \textit{E. coli}~\citep{turnerReplayingEvolutionTest2015}, Phage $\lambda$ infection of \textit{E. coli}~\citep{meyerRepeatabilityContingencyEvolution2012, guptaHostparasiteCoevolutionPromotes2022a}, colistin resistance~\citep{jochumsenEvolutionAntimicrobialPeptide2016a} and diversity~\citep{al-tameemiMicrobialDiversificationMaintained2024} in \textit{Pseudomonas} species, and genetic interactions in yeast \citep{vignognaExploringLocalGenetic2021}.
Recently, replay experiments have expanded beyond bacteria, bacteriophage, and fungi to study mitochondria evolution in nematodes \citep{dubieDissectingSequentialEvolution2024}.
See \citep{blountContingencyDeterminismEvolution2018} for a review on analytical replay experiments, including comparisons to other methods for analyzing historical contingency in evolution. 

Beyond laboratory systems, analytical replay experiments have also been used in digital evolution experiments as an empirical tool for quantifying the role of past events in the evolution of novel traits. 
For example, replay studies using the Avida Digital Evolution Platform~\citep{ofria_avida_2009} have investigated re-evolution following a near-extinction event \citep{yedidHistoricalContingentFactors2008}, the role of deleterious mutations in long-term evolution \citep{covertiiiExperimentsRoleDeleterious2013}, and the potentiation of associative learning \citep{fergusonPotentiatingMutationsFacilitate2023}. 
Outside of Avida, analytical replay experiments have been used to study developmental exaptations \citep{rennerComputationalModelDevelopmental2024} and the increased evolutionary exploration experienced during a selective sweep \citep{fergusonPredictingUnpredictableUsing2024}.

This work provides an introduction to analytical replay methods for the evolutionary computing community.
Most applications of replay experiments in laboratory and \textit{in silico} evolution experiments focus on identifying mutations that potentiate the evolution of novel traits. 
Likewise, we can use replay experiments in an EC context to illuminate events that potentiate problem-solving success. 
For example, what forms of variation correlate most with increases in success potentiation? 

Analytical replay experiments have broader applications in EC beyond studying which mutational steps potentiated success. 
For example, we could design replay experiments to investigate how particular parent selection decisions influenced success, or we could disentangle which events during an evolutionary search potentiated a problematic collapse in candidate solution diversity. 
Such studies can deepen our understanding of why evolutionary searches succeed or fail and illuminate why particular algorithm design choices produce different evolutionary dynamics, such as premature convergence, diversity maintenance, or bloat. 

In the remainder of this chapter, 
we overview how to design replay experiments and discuss their limitations (Section~\ref{sec:replay-experiments}). 
We then present an example replay experiment in a genetic programming context, stepping through design decisions, results, and additional considerations (Section~\ref{sec:demonstrations}). %
Finally, we conclude with a brief discussion of promising future research directions using replay experiments (Section~\ref{sec:conclusion}).

\section{Analytical replay experiments}
\label{sec:replay-experiments}

Analytical replay experiments allow us to empirically measure how long-term evolutionary outcomes are contingent on a population's accumulated history. 
Specifically, replay experiments sample the distribution of possible outcomes if we could ``rewind'' time and restart a population's evolution at a series of historical time points, each representing a different amount of accumulated history.
Comparing evolutionary changes in the original, historical population to the distributions of outcomes observed in replays can reveal when critical potentiating events occurred in the historical population. 

In both laboratory and \textit{in silico} systems, replay experiments can be prohibitively costly in terms of storage and time. %
To identify all generations with potentiating events, an experimenter would need to replay the historical population's evolution at each generation. 
Such high-resolution replays would require (1) snapshots (or samples) of the historical population at each generation to initialize replays from and (2) running many replicate replays from each of those snapshots to observe a representative distribution of possible outcomes.
Moreover, the experimenter would need to record detailed information about events that occurred in the historical population in order to disentangle which particular historical events correlate with evolutionary outcomes of interest.  

Fortunately, approaching replay experiment design with a hypothesis-driven mindset can help to maintain tractability by informing what data should be collected and at what resolution. 
With this in mind, we recommend selecting a well-defined evolutionary outcome of interest in an evolved population, such as problem-solving success. 
Then, select a focal event type hypothesized to have potentiated the outcome of interest, such as mutation or recombination events.
These choices can then guide further experimental design. 
In the remainder of this section, we provide a guide for designing replay experiments, including critical considerations for experimental design and analysis decisions.

\subsection{Designing and conducting replay experiments}
\label{sec:replay-experiments:design}

\begin{figure}[h]
    \centering
    \includegraphics[width=0.99\linewidth]{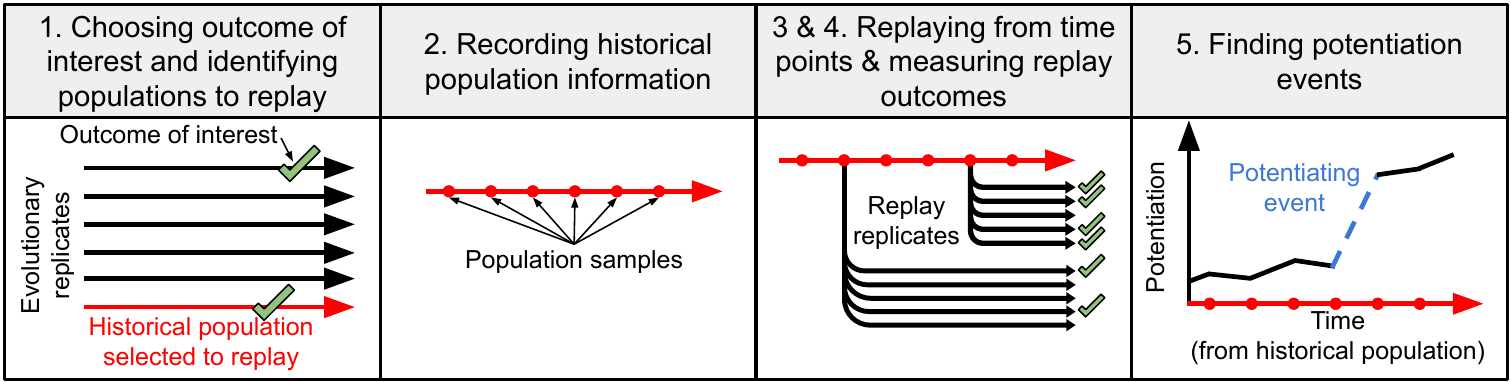}
    \caption{
        \textbf{Conceptual figure demonstrating the five steps of designing and running replay experiments.}
        The first pane shows the evolution of independent replicates, two of which evolved our outcome of interest (green checkmarks). 
        In this example, the bottom replicate was chosen to be our historical population for replays (red throughout). 
        The second pane shows population samples collected from the historical population at multiple time points. 
        The third pane shows steps 3 and 4: we replay replicates founded from two time points. 
        Replay replicates from the later time point evolve the outcome of interest more frequently, indicating greater potential for the outcome of interest. 
        Finally, we collate the replay results into a potentiation over time plot, where potentiation is calculated as the percentage of replay replicates founded from that time point that evolve the outcome of interest. 
        In this hypothetical example, we have identified a potentiating event in the  timestep shown in dashed blue.  
    }
    \label{fig:design:conceptual}
\end{figure}

We divide replay experiment design and execution into five key steps (depicted in Figure~\ref{fig:design:conceptual}):

\begin{enumerate}
    \item \textbf{Choosing} an outcome of interest and identifying populations to replay
    \item \textbf{Recording} population information at time points of interest to start replays from
    \item \textbf{Replaying} evolution by evolving ``replay replicates'' at each selected time point
    \item \textbf{Measuring} the outcomes of each replay replicate
    \item \textbf{Analyzing} trends in replay outcomes over time to identify potentiating events
\end{enumerate}

\subsubsection{Step 1: Choosing an outcome of interest and identifying populations to replay}
\label{sec:replay-experiments:design:choosing-an-outcome}

The first step in designing a replay experiment is to identify an evolved population to replay.
Choice of population is tightly linked to which evolutionary outcomes the practitioner aims to investigate, and that choice often occurs under one of two different scenarios: 
(1) \textit{opportunistic selection} of a population evolved \textit{a priori} for a different purpose 
and (2) \textit{crafted selection} from among populations evolved for the purpose of replaying their evolution. 

Early replay experiments were used to identify mutations that potentiated the evolution of a rare trait, such as citrate metabolism in \textit{E. coli}~\citep{blountHistoricalContingencyEvolution2008}. 
In these experiments, Blount \textit{et al.} chose which populations to replay opportunistically.
That is, a set of historical populations had been evolved as part of the Long-term Evolution Experiment~\citep{lenski_convergence_2017} without replay experiments planned \textit{a priori}. 
In this example, experimenters observed a surprising outcome of interest (the evolution of citrate metabolism) in an evolving population, and then decided to use replay experiments to identify potentiating mutations that led to the evolution of a rare trait. 

In EC, evolution frequently produces surprises~\citep{lehmanSurprisingCreativityDigital2020}, and replay experiments can be used as an opportunistic tool for more deeply understanding what led to a surprising outcome in a population.
Applying replay experiments opportunistically in this way requires detailed record keeping of the original population's history, which may not have been collected and saved.
However, EC systems are often advantaged in this respect, as practitioners can exactly reproduce a surprising result with additional data collection instrumentation if their evolutionary system allows for perfect replication for a given random number seed.

Historical populations can also be evolved for the purpose of subsequent replay experiments, allowing experimenters to craft population selection to precisely target a particular research question or hypothesis. 
For example, Ferguson \textit{et al.}~\citep{fergusonPredictingUnpredictableUsing2024} used replay experiments to investigate how selective sweeps can enhance the likelihood of a population crossing a valley of deleterious mutations by temporarily weakening selection pressure on individuals at the leading edge of the sweep. 
Ferguson \textit{et al.} evolved initial populations with this research question in mind, allowing them to target a specific evolutionary outcome of interest (crossing fitness valleys). %

In the context of evolutionary computing, we are often interested in outcomes related to our capacity to discover solutions to a problem, such as: 
final problem-solving success or failure, 
the evolution of a phenotypic or behavioral characteristic of interest,
solution robustness, 
improvements on a particular objective for a multi-objective problem,
increases or decreases in population diversity,
premature convergence, or
changes in model complexity and/or bloat.
Any outcome of interest observed in an evolved population can be selected as the focus of a replay experiment as long as that outcome is sufficiently well-defined to allow for its detection and/or quantification in other populations. 

Replay experiments are most informative for understanding \textit{rare} evolutionary outcomes.
For example, imagine that we evolve 50 populations, each starting from identical initial populations, and evolution of each of the 50 populations produces a valid solution to the problem at hand.
In this case, the initial population is already fully potentiated for problem-solving success in the context of the evolutionary system in use. 
That is, the outcome (finding a solution) is not contingent on any rare sequence of events specific to any particular population's history, and replay experiments would be of little value.
In contrast, imagine that we evolve 50 replicate populations, each starting from an identical initial population, and the outcome of interest occurs in only 1 of the 50 replicates. 
In this scenario, the outcome of interest had low potentiation in the starting population; that is, it had a low initial probability of occurrence. 
In the replicate where the outcome of interest was observed, there is likely a rare sequence of events that potentiated (\textit{i.e.}, increased the likelihood of) that outcome.
We can often identify the actualizing event that led directly to the outcome of interest by analyzing the events that occurred immediately prior to the outcome, such as the final mutation that transforms a non-solution into a solution. 
Replay experiments provide a empirical tool for deeper understanding: what sequence of events in a population's history increased the likelihood of the actualizing event?  

\subsubsection{Step 2: Recording population information for replays}
\label{sec:replay-experiments:design:recording-population-information}

To conduct replay experiments of a chosen population, we initialize additional evolutionary replicates founded from different time points along that population's evolutionary history. 
Laboratory experiments have limited options for recording evolutionary history: researchers typically freeze samples from their populations at regular intervals, allowing for future replays starting from each frozen sample. 
Fortunately, computational studies have greater flexibility, with multiple methods for recording history as described below. 
Regardless of what method you use to sample a population, tracking and saving additional population data during a run can be costly in terms of time and storage. 
To combat these bottlenecks, we suggest running initial evolutionary replicates \textit{without} the additional record keeping. 
After identifying replicates for replays, many EC systems can then perfectly rerun a replicate using the same random number seed, and additional data tracking can be enabled without altering the evolutionary results.

In many EC systems, we can  save the complete state of the population at any point in time (e.g., as done in~\citep{fergusonPredictingUnpredictableUsing2024, rennerComputationalModelDevelopmental2024}).
This snapshot can contain all information needed to recapitulate the state of the population, including the genomes of all individuals in the population and any additional environmental or contextual information.
For example, if using a spatially structured population or island model, you should consider initializing replay replicates with genomes in historically accurate positions. 
Starting replays from full snapshots perfectly encapsulates historical population dynamics, offering the most detailed view of how accumulated history altered evolutionary outcomes. 

In systems where perfect snapshots are not possible or too large to tractably store, we can sample the population at time points of interest. 
In the extreme case, sampling a single representative genotype (e.g., most abundant or maximum fitness) from a population can still lead to informative replays. 
However, smaller samples may be less representative of the population, especially if there are critical interactions between individuals in the population or important ecological dynamics that have strong influences over evolution. 
For example, many diversity-preserving selection algorithms (e.g., fitness sharing, lexicase selection, etc.) create implicit ecological dynamics within a population of candidate solutions~\citep{dolson_ecological_2018}.
Systems where candidate solutions are evaluated against one another (e.g., coevolutionary systems or when evolving game playing agents) may also require large samples or full snapshots for the outcomes of replay experiments to be representative. 
Sampling methods have been shown to bias replay results in consistent ways \citep{escondo_summarizing_2026}, and thus we recommend using full snapshots for replay experiments when possible. 

Another common technique for recording representative population information is to track phylogenetic information (i.e., parent-offspring connections) over time, pruning branches in the phylogeny as they go extinct.
This approach works best for populations evolved with asexual reproduction (i.e., without crossover or other recombination operators); otherwise, efficient pruning becomes more challenging. 
If phylogenetic information is available, a representative genotype at the end of an evolutionary search can be selected (e.g., an evolved solution, the most abundant genotype, \textit{etc.}), and that genotype's lineage can be traced back to the original starting population. 
We can then replay evolution from different genotypes along this lineage as done in~\citep{fergusonPotentiatingMutationsFacilitate2023}.
See~\citep{dolson_interpreting_2020} for more information on phylogeny tracking. %

\subsubsection{Step 3: Replaying from selected time points}
\label{sec:replay-experiments:design:replaying-from-selected-time-points}

When starting new replay replicates from a population's history, there are two questions to consider. 
First, what time points should we replay?
And second, how should we parameterize the replay replicates?

\vspace{0.1in} \noindent \textbf{Selecting historical time points to replay:}
A population's long-term potential for a focal outcome is not guaranteed to increase monotonically over time (e.g., see \citep{fergusonPotentiatingMutationsFacilitate2023, fergusonPredictingUnpredictableUsing2024, rennerComputationalModelDevelopmental2024}). 
When analyzing replay experiments, we typically screen for large changes in potentiation within in a narrow time window (e.g., a single generation), as they indicate when critical events happened in a population's history. 
However, we cannot be guaranteed to identify exactly when potentiating events occur without replaying a population's evolution from each generation.  
For populations with a shallow evolutionary history evolved in a computationally efficient system, such exhaustive replay experiments are tractable (e.g., see~\citep{fergusonPredictingUnpredictableUsing2024,rennerComputationalModelDevelopmental2024}).
For populations with longer histories evolved in more complex systems, we recommend replaying evolution from a limited number of past time points in the focal population's history as is done in laboratory replay experiments~\citep{blountHistoricalContingencyEvolution2008}. 
Even with this lower resolution approach, we can still effectively screen for precise events that triggered changes in potentiation for a particular evolutionary outcome using a ``divide and conquer'' approach similar to a binary search algorithm.
Begin with performing distantly spaced replays along a population's history.
Then, screen for large changes in potentiation using this lower resolution replay data, and perform additional, higher-resolution replays in windows in which a large change in potentiation occurred. 
This process can be repeated until achieving a sufficiently high resolution understanding of when an outcome of interest was potentiated; see~\citep{fergusonPotentiatingMutationsFacilitate2023} for an example of this approach. 

\vspace{0.1in} \noindent \textbf{Parameterizing replay replicates:}
After choosing when to replay a population's evolution, we must decide how to parameterize those replays. 
Replay replicates typically keep all experimental parameters (e.g., population size, mutation rate, \textit{etc.}) exactly the same as the original population's evolutionary replicate.
However, each replay replicate uses a unique random number seed to ensure different sequences of stochastic events across replays. 
Due to time constraints, laboratory replay experiments evolve all replay replicates for the same number of generations; for example, all replays in \citep{blountHistoricalContingencyEvolution2008} evolve for roughly 3,700 generations. 
In computational systems, we recommend configuring each replay replicate to run for the same number of \textit{total} generations as the historical population being replayed. 
For example, imagine that our original historical population evolved for 1,000 generations.
Replay replicates founded from generation 300 of the historical population would be evolved for 700 additional generations, while replays founded from generation 450 would be evolved for 550 generations. 
In this way, all populations (historical and replays) undergo the same total number of generations of evolution.

\vspace{0.1in} \noindent \textbf{Engineered replays:}
Conventional replay experiments initialize replay replicates from states drawn from the focal population's history. 
We term any replay initialized from an \textit{ahistorical} population state or parameterized differently than the historical population as an ``engineered replay.''
Cleverly designed engineered replays can further isolate the specific events driving changes in a population's potential in situations where a conventional replay cannot.

Consider an example where we replay evolution starting from two adjacent genotypes along a lineage (i.e., one ``parent'' genotype and one ``offspring'' genotype), and these two genotypes differ by two mutations (referred to as mutation A and mutation B).
Using a conventional replay experimental design, we would run two sets of replay replicates, each starting from one of these two genotypes: the parent genotype with neither mutation and the offspring genotype with both mutations A and B. 
Differences in the distribution of outcomes between these two sets of replays would reveal the influence of those two mutations overall but not each mutation's individual importance.
That is, we would not know whether just one of the two mutations drove a change in potential or if it was the co-occurrence of the mutations together that drove any changes in potential.  
To disentangle the individual effects of these mutations, we can engineer new populations with different combinations of mutations \textit{as they could have been} and initialize replays with those engineered populations. 
In this example, we might run four sets of replays: one set from the parent genotype, one set of the offspring genotype as it appears in the lineage (with both mutations A and B), one set with an engineered offspring genotype with just mutation A, and one set with an engineered offspring genotype with just mutation B. 
The two engineered genotypes may not represent actual historical states in the original evolved lineage, but they help us to disentangle which specific differences between two focal time points drove any observed changes in potential.

The exact form that an engineered replay takes will depend on the specific hypothesis or research question. 
Possibilities include artificially altering the makeup or diversity of a historical population, altering the replay environment or fitness measure in some way, altering the spatial arrangement of genotypes in the population \citep{fergusonPredictingUnpredictableUsing2024}, altering the selection procedure, \textit{et cetera}.
We strongly recommend conducting engineered replays in tandem with conventional replays of the original population's evolution to provide a baseline comparison. %

\subsubsection{Step 4: Measuring outcomes of each replay replicate}

Most replay experiments have examined a population's potential for evolving a focal trait, using replays to examine how that potential changed over the population's history. 
In this context, replay replicates can be categorized as ``successful'' or ``unsuccessful'' based on whether they evolved the focal trait.
We use the percentage of successful replay replicates founded from a time point as the \textit{potentiation} of the population at that point in time \citep{blountHistoricalContingencyEvolution2008, turnerReplayingEvolutionTest2015, meyerRepeatabilityContingencyEvolution2012, fergusonPotentiatingMutationsFacilitate2023, fergusonPredictingUnpredictableUsing2024, rennerComputationalModelDevelopmental2024}. 
In the context of evolutionary computing, problem-solving success can be treated exactly as focal traits have been treated in prior laboratory and \textit{in silico} evolution studies. 

Replay experiments are not limited to measuring potentiation of a focal trait.
For example, we can also measure the final fitness distribution of replay replicates as has been done in laboratory~\citep{woodsSecondorderSelectionEvolvability2011} and \textit{in silico}~\citep{sansonComputationalBaselineChanges2025} studies. 
The final distribution of fitnesses observed from a set of replay replicates sample the long-term fitness distribution of a population from a given point in time, creating a measure of evolvability in that population. 
Likewise, we could measure the accumulation of phenotypic, genetic, or phylogenetic diversity across replay replicates to measure a population's \textit{potential} for generating and maintaining diversity.
In genetic programming, we might be interested in measuring the distribution of bloat or model complexity across replays to screen for bloat potential over a population's history.
Even if a single trait or problem-solving success is the main focus of a replay experiment, these additional distributions can be collected during the same replay experiment, providing a more complete picture of population's potential at different time points.

\subsubsection{Step 5: Finding potentiating events}

Quantifying a population's potential to evolve a trait of interest can be informative on its own. 
However, replay experiments can also identify \textit{potentiating events} in a population's history. 
When we consider evolution of a focal trait or problem-solving success, the \textit{actualizing event} is the mutation that confers the trait or final success. 
In the case of the evolution of citrate metabolism in \textit{E. coli}~\citep{blountHistoricalContingencyEvolution2008}, the actualizing event was the mutation that conferred the ability to metabolize citrate.
When the local mutational neighborhood of a population contains a highly beneficial trait like citrate metabolism, that trait is likely to be selected for if the actualizing mutation occurs. 
Replay experiments can identify the \textit{potentiating events}; that is, identify the events in a population's history that made the focal trait more likely to evolve. 
For example, Ferguson \textit{et al.}~\citep{fergusonPotentiatingMutationsFacilitate2023} used replay experiments to isolate single mutations that potentiated the evolution of associative learning, with one experiment identifying a single deleterious mutation that increased the likelihood of evolving associative learning by 64 percentage points, with the potentiating event occurring 26 lineage steps before the actualizing mutation that conferred associative learning. 

To identify potentiating events, we must perform replays from a series of time points in the original population's history. 
When the likelihood of an outcome of interest changes between two sequential replay start points, we know that a potentiating event occurred.
However, replay experiments alone do not identify what caused (or influenced) the change in potentiation, especially if there are large windows of time between successive replay start times. 
As previously discussed (Section~\ref{sec:replay-experiments:design:replaying-from-selected-time-points}), we can start with low resolution screens for potentiating events and iteratively increase the resolution around large changes in potential to identify precise times where potential meaningfully changed. 
Once we know exactly when an important potentiating event occurred, we can try to identify the event or set of events underpinning the change in potential. 

What do potentiating events look like? 
To the best of our knowledge, all analytical replay experiments in the laboratory and \textit{in silico} studied asexual populations, and screened for potentiating mutations. 
Indeed, isolating critical mutational changes most directly reveals the step-by-step potentiation of an evolved trait. 
For many EC systems, practitioners should additionally screen for potentiating recombination operations as important sources of genetic variation. 

Because most replay experiments focus on mutationally driven changes in evolutionary potential, the influence of population dynamics on a population's evolutionary potential is understudied using replay experiments. 
Indeed, selection events or the spatial distribution of organisms can alter the potential of the population \citep{fergusonPredictingUnpredictableUsing2024}.
Ecological interactions or environmental effects, such as a change in temperature or a limitation on a resources, could also serve as potentiating events.
Choosing what types of potentiating events to screen for will depend on the particular outcome of interest and research objectives, but we do encourage practitioners to think beyond mutationally-driven potentiation in future studies.

Recording more detailed data about a broad range of events that occurred during the historical population's evolution allows for a more comprehensive study of which particular events could have driven a change in potentiation. 
Analyzing events at a focal time step involves practitioner intuition about a system: looking at what changes from one time point to the next, and making informed hypotheses about what events could have driven those changes. 
Experimental manipulations (e.g., follow-up engineered replays) will often be necessary to confidently differentiate between potentiating and non-potentiating events. 

Instead of focusing on specific potentiating events, we can study what kinds of events in a population have strong correlations with changes in potential across multiple sets of replays. 
In an EC context, this could reveal if the type of genetic variation produced by particular mutation or recombination operators tends to increase or decrease a population's evolutionary potential more often than expected by chance.
Such studies could inform future improvements to an EC system. 

\subsection{Limitations of replay experiments}

While replay experiments are a powerful tool for measuring a population's evolutionary potential, several factors can limit their applicability.  
Here, we outline critical limitations to replay experiments, and we discuss how they can be addressed as well as how they might change in the future.

\subsubsection{All replays are retrospective}

True analytical replay experiments (\textit{sensu} \citep{blountContingencyDeterminismEvolution2018}) replay an \textit{existing} population's evolution.
This creates two limitations inherent to replay experiments: (1) replay experiments are not intended to be used at runtime to improve evolutionary search decisions, and (2) replay experiments are limited by our historical populations.

Replay experiments allow us to measure a genotype or population's long-term evolutionary potential, and so it is tempting to design evolutionary search algorithms that incorporate this information into their selection decisions at runtime. 
However, using replay experiments for runtime decisions is fundamentally intractable: running the replay experiment to quantify potential would take \textit{longer} than simply allowing the original evolutionary run to finish. 
Replay experiments help us to understand \textit{why} or \textit{how} an outcome occurred post-hoc. 
While replay experiments are not directly helpful for improving search performance at runtime, we argue that insights gained from retroactive replay experiments can help us to improve existing evolutionary search methods or design new methods entirely. 

Consider a situation where our EC system struggles to produce a solution to a difficult problem. 
If we have \textit{any} successful evolutionary replicates, we can run replays experiments on them to uncover the events that potentiated their success. 
This information could help us understand why our EC system rarely solved the problem, and inform how we approach future problems. 
However, replay experiments are less useful if we have yet to observe the evolution of an outcome of interest (e.g., problem-solving success). 
Replays screen for changes in evolutionary potential by comparing distributions of outcomes from replay replicates started at different historical time points. 
If the distribution of outcomes from replays at all time points is identical (e.g., problem-solving failure), the population's potential will be measured as unchanged for its entire history. 
In the case of total problem-solving failure across all evolutionary replicates, it may be tempting to replay populations that achieved the largest fitness score to measure potential for higher fitness; however, until we have a better understanding of how evolutionary potential changes over time, we cannot necessarily assume potential for higher fitness (relative to other replicates) would be indicative of potential for future problem-solving success.

\subsubsection{Are replay-based insights generalizable?}

Imagine that we use replay experiments to screen for events that potentiate success for a particular evolutionary replicate. 
Can we then improve the success rate of future replicates by adjusting our EC system to make the potentiating events more likely? 
More broadly, do potentiation insights from one population generalize to other populations? 
And, do potentiation insights from one evolutionary system generalize to other systems? 
Are some types of insights generalizable while others are not? 
These questions are currently unanswered. %

Because laboratory and computational replay experiments are increasing in popularity and scale, performing the meta-analyses necessary to answer questions about generalizability is becoming feasible. 
For example, does potentiation increase gradually or in short bursts? 
What are common lag times between potentiating events and their actualizing events? 
How often does potentiation decrease before the target trait evolves?
We suspect future work will soon begin answering these questions.

\subsubsection{Replays are computationally expensive}

While powerful, evolutionary search is an inherently costly means of problem solving, as it requires many repeated evaluations of candidate solutions. 
Replaying a single population requires many additional evolutionary replicates, each founded from different time points in the original population's history. 
In evolutionary systems that require substantial compute to evolve a single population, replay experiments are likely intractable. 
Replay experiments may also require an intractable amount of storage in systems with very large population sizes, genomes that cannot be easily compressed, or large amounts of contextual information required to accurately restart a population's evolution.

The costs of replay experiments can be ameliorated by lowering their resolution in two ways.
To reduce the runtime costs of replay experiments, we can replay evolution from a smaller set of historical time points, screening for windows of time where further replay might be especially valuable (as described in Section \ref{sec:replay-experiments:design:replaying-from-selected-time-points}). 
However, this approach may miss ephemeral changes in potentiation, such as a drastic increase followed by a similarly sized decrease. 
To reduce the storage requirements of replay experiments, we can use small samples of a historical population to initialize replay replicates from instead of using full snapshots (as described in Section~\ref{sec:replay-experiments:design:recording-population-information}), though sampling has been shown to bias replay results \citep{escondo_summarizing_2026}.

\section{Demonstrating replay experiments with a genetic programming system}
\label{sec:demonstrations}

Here, we provide a demonstration of designing and running replay experiments in an EC context. 
To illustrate how to conduct replay experiments, we organize this section as a step-by-step discussion of running an example replay experiment from the experimenter's point of view, mirroring the guide given in Section~\ref{sec:replay-experiments:design}.

\subsection{Step 1: System design, choosing an outcome, and identifying populations to replay}

In this demonstration, our goal is to highlight the utility of replay experiments in EC. 
As such, we evolved populations with the intent of replaying them (i.e., crafted population selection). 
We selected our outcome of interest to be perfect problem-solving success. 
Thus, our replay experiments measure the historical population's potential for evolving a perfect solution and how that potential changed over time. 

Typically, potentiation studies are conducted on rare outcomes, so we designed our demonstration system such that perfect solutions rarely evolved. 
Using a simple genetic programming (GP) system, we evolved populations of 500 linear genetic programs for 200 generations on the ``small or large'' program synthesis benchmark problem~\citep{helmuthGeneralProgramSynthesis2015}.
To solve the small or large problem, a program must categorize an integer input as either small ($< 1000$), large ($> 2000$), or neither. 
Each generation, we evaluated programs in the population on 100 pass/fail training cases, and we used the lexicase selection algorithm~\citep{helmuthSolvingUncompromisingProblems2015} to choose parents to produce offspring.
Parent programs produced offspring asexually, applying the following variation operators to offspring: 
single-instruction substitutions ($0.2\%$ per-instruction),
argument substitutions ($0.2\%$ per-argument),
single-instruction insertions and deletions ($0.2\%$ per-instruction),
and slip mutations, which can duplicate or delete sequences of instructions~\citep{lalejini_gene_2017} ($5\%$ per-program). 
We limited program length to 128 instructions, and when evaluating a program on a test case, we gave the program 128 time steps (i.e., instruction-execution steps) to return its output.
A program is categorized as a solution if it solves the complete training set used during selection as well as a set of 1,000 testing cases not used during evolution. 
We did not tune this GP system for solving this problem, and we ran evolution for relatively few generations to decrease the computational cost of these replay experiments. 
Our code, configuration files, and analyses are available in the supplemental material~\citep{supplement}.

\begin{figure}
    \centering
    \includegraphics[width=0.75\linewidth]{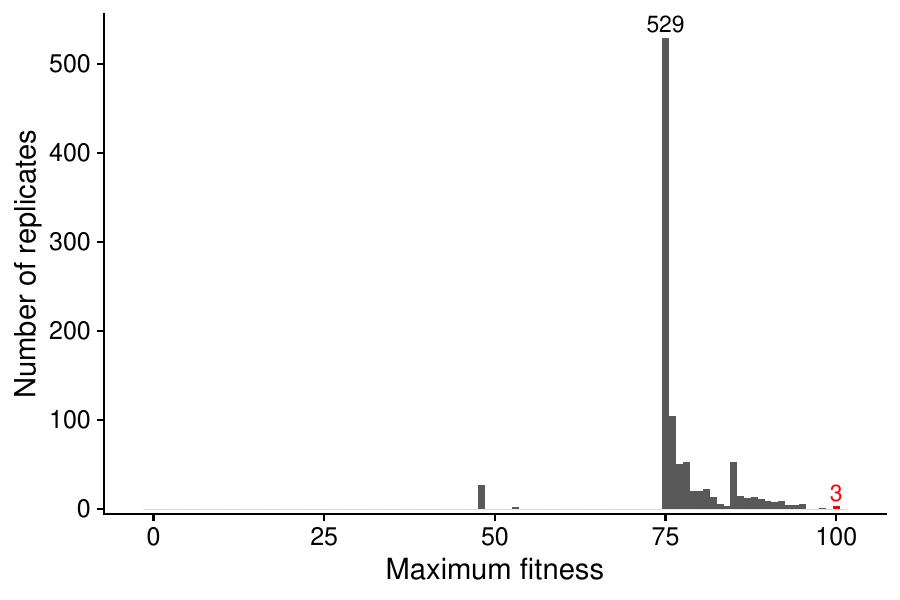}
    \caption{
    \textbf{Histogram (bin width = 1) showing the distribution of maximum fitness reached by all 1,000 initial evolutionary replicates.}
    Fitness was in the range $[0\%,100\%]$. 
    Three replicates (out of 1,000) reached the maximum fitness, while the majority of replicates reached exactly 75\% fitness (labels shown). 
    }
    \label{fig:demo:basic_potentiation_initial_fitness}
\end{figure}

To identify populations to replay, we evolved 1,000 independent populations, each initialized with random candidate solutions.
We define a candidate solution's fitness as the percentage of training cases that it solves ($[0\%,100\%]$). 
The majority of replicates (529 of 1,000) evolved a maximum fitness of 75\%, indicating a local optimum that frequently traps populations (Figure \ref{fig:demo:basic_potentiation_initial_fitness}).
Of the 1,000 replicates, three evolved solutions to the problem, indicating that a population starting from a randomly generated ancestor has a 0.3\% probability of evolving a perfect solution in this system. 
For this work, we selected the three replicates that evolved solutions for additional replay analyses.
We will refer to these historical populations as Populations A, B, and C. 
While outside the scope of this demonstration, replaying the lowest-fitness replicates or a sample of replicates stuck at the 75\% fitness local optimum may also provide insight on \textit{why} these replays failed to evolve a solution.

\subsection{Step 2: Recording population information for replays}

In our initial runs, we did not store detailed historical records to reduce our memory storage footprint.
Once we chose our three populations to replay, we perfectly reproduced them with identical random number seeds, except we enabled comprehensive data tracking to use for replays in the next step. 
For this demonstration experiment, we maximized our resolution for detecting changes in problem-solving potential by recording full population snapshots at every generation until a solution evolved.

\subsection{Steps 3 and 4: Running replay replicates and measuring outcomes}

For each of our three chosen historical populations, we replayed evolution from each generation up until a perfect solution evolved in the historical population (generations 75, 120, and 172, respectively). 
For each generation, we evolved 100 replay replicates. 
We configured each replay replicate with a unique random number seed, founding each replicate from the full population snapshot taken at that generation.
All replays were given the same number of \textit{total} generations (200) as the historical population (e.g., we evolved replays replicates founded from generation 50 for 150 generations). 
All other parameters for the replay replicates remained identical to those used to evolve the historical population. 
We recorded whether each replay replicate successfully evolved a perfect solution. 
For each generation of the original population, we calculated the potential to evolve a perfect solution \textit{from that state} as the percentage of successful replay replicates founded from that generation. 

\subsection{Step 5: Analysis of replay results and identification of potentiating events}

\begin{figure}[h]
    \centering
    \includegraphics[width=0.95\linewidth]{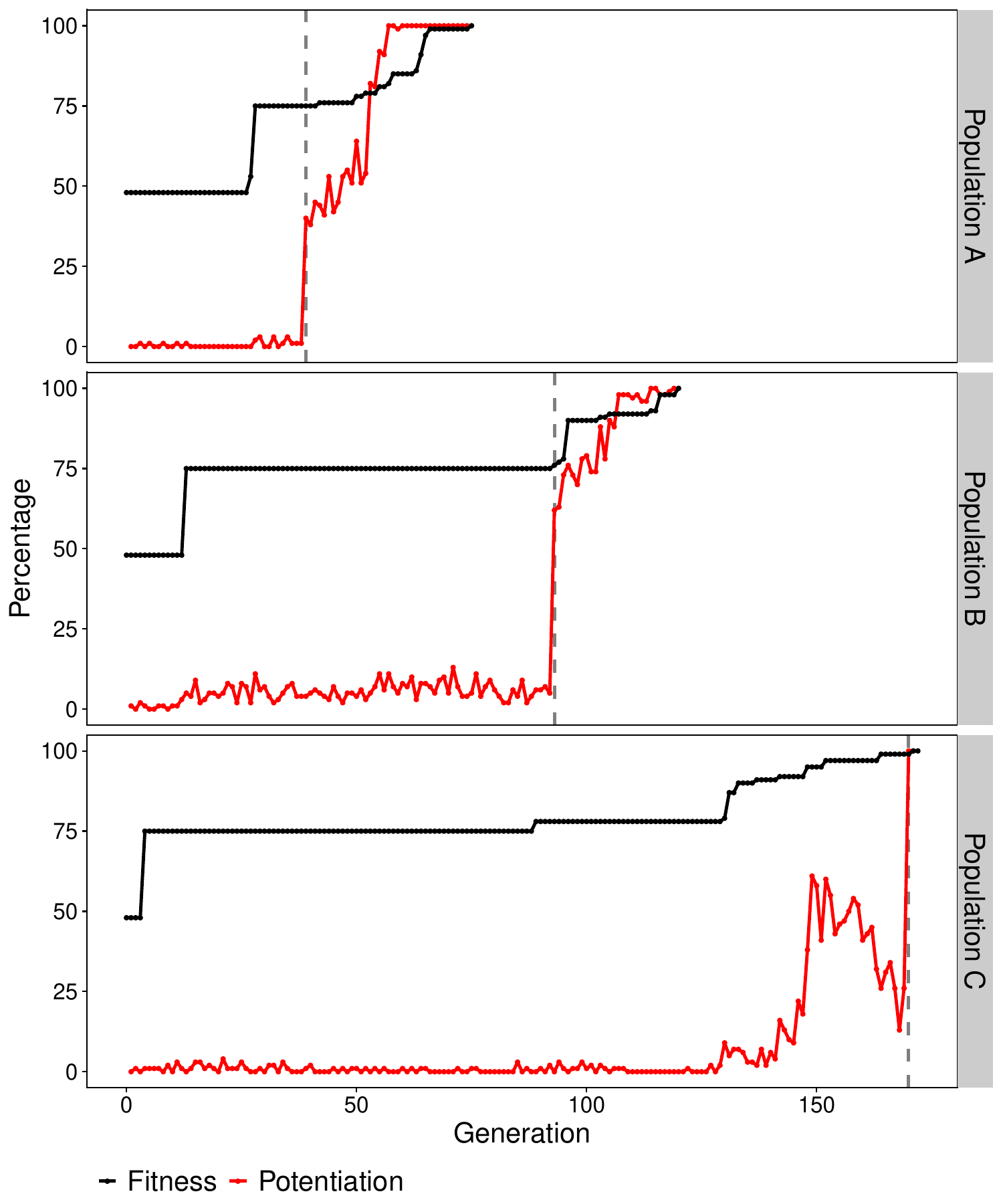}
    \caption{
    \textbf{Maximum fitness and potentiation over time for the three historical population replayed.}
    Fitness is shown as the percentage of training cases passed by the single best solution in the population. 
    Potentiation is calculated as the percentage of replay replicates founded from that generation that successfully evolved a perfect solution. 
    Both lines stop at the generation a perfect solution evolved.
    For each population, the dashed vertical line shows the generation with the largest increase in potentiation. %
    }
    \label{fig:demo:basic_potentiation_rep_1}
\end{figure}

Figure~\ref{fig:demo:basic_potentiation_rep_1} shows fitness and potentiation over time for the three historical populations replayed.
Examining Population A, maximum fitness in the historical population stays consistent at 48\% until a jump to 75\% at generation 28. 
This fitness increase does not coincide with a substantial increase in potentiation. 
Instead, potentiation of the historical population rises sharply at generation 39, jumping from 1\% to 40\%. 
This jump in potentiation does not correspond with a change in the maximum fitness in the population; the population stays at the local optimum of 75\% fitness. 
From there, maximum fitness and potentiation increase intermittently, with the population becoming effectively fully potentiated at generation 57, 18 generations before a perfect solution evolved. 
Note that replay experiments produce an \textit{estimate} of potentiation, so a result of 0\% or 100\% potentiation may not be a true 0\% or 100\%.  
This first replayed population demonstrates that potentiating events may not correspond with increases in fitness. %
The potentiating event occurring between generation 38 and 39 made problem-solving success substantially more likely, but the maximum fitness in the population remained unchanged. 
Indeed, we would not be able to detect this potentiating event without the use of replay experiments.

Maximum fitness of the Population B shows a similar trend to Population A: an initial maximum fitness of 48\%, followed by a period of 75\% fitness, and then small increases to a perfect solution at generation 120. 
Early potentiation was noisy, fluctuating between 0\% and 13\% potentiation. 
Potentiation for success dramatically increased at generation 93 from 5\% to 62\%. 
Unlike the first historical population, here we see the increase in potentiation coinciding with an increase in fitness; the maximum fitness in generation 93 increased from the local optimum of 75\% to 76\%.

Finally, Population C evolves past the local optimum of 75\% fitness without potentiation increasing.
Of the three populations replayed, Population C showcases that potentiation can substantially \textit{decrease} as the population evolves. 
Much like Population B, we see the population evolve to a local optimum (99\% fitness) before an event during generation 170 both potentiates and actualizes a perfect solution. 

While we have identified the \textit{generations} that potentiated these populations, what were the actual \textit{potentiating events}?
As a first pass, we again reproduced our three historical populations, recording the phylogeny tracing from all extant organisms back to the original ancestors. 
Examining the lineage at the potentiating time points, the dominant lineage in Population B experienced two mutations (one substitution and one insertion) while Population C experienced two substitution mutations. 
Interestingly, Population A does not have any changes along the dominant lineage at the potentiating time point. 
To narrow the potentiating events down further, we could run engineered replay experiments. 
For example, for Population B we could identify the potentiating mutation by running two additional replay replicates: we would split the two mutations, running one set of replay replicates with just the first mutation and a second set of replicates with only the second mutation.

Why do we observe different potentiation dynamics between these three populations? 
By replaying three populations, we have only scratched the surface of the potentiation dynamics of this system. 
While further replays fall outside the scope of this chapter, we hope the replays in this simple system demonstrate that replay experiments are a powerful tool for investigating long-term evolutionary dynamics, but we, as a field, still need to build our intuition around replay experiments and evolutionary potential.

\section{Conclusion}
\label{sec:conclusion}

In this chapter, we provide an introduction to analytical replay experiments in the context of evolutionary computing.
Laboratory and computational replay experiments have been a powerful analytical tool for understanding how events in a population's history have contributed to particular evolutionary outcomes. 
We argue that replay experiments can also be a valuable tool for more deeply understanding evolutionary search algorithms, serving as an additional approach to strengthening the theoretical foundations of evolutionary computing. 

Replay experiments enable many promising research directions. 
Broadly, the dynamics of potentiation for different kinds of evolutionary outcomes remain understudied, and computational evolution systems can be tractable systems for testing hypotheses that would otherwise be challenging to investigate in laboratory systems. 
In EC, replay experiments can be applied to studying problem-solving success as we demonstrated in this work. 
What kinds of events correlate with changes in potentiation for problem-solving success or failure?
Do these events match the EC community's collective intuition, and does this information help us improve existing evolutionary search algorithms or inspire new algorithms altogether? 
For a given EC system, do some forms of genetic variation have stronger correlations (positive or negative) with changes in potentiation more often than other forms of genetic variation? 
If so, how do these correlations change across problems, genetic representations, and system configurations? 

We can also study evolutionary potentiation for outcomes other than problem-solving success, such as diversity, solution bloat or complexification, and evolvability. 
Can we identify events in a population's history that potentiated (or suppressed) collapses in diversity or premature convergence? 
How much are outcomes related to diversity driven by historical contingency versus inherent to particular evolutionary algorithms? 
If there are particular events that signal unwanted diversity collapses, can we detect and counteract them at runtime?

For many focal outcomes, we can use replay experiments to construct potentiation landscapes.
A fitness landscape maps genotypes to their corresponding fitness value.
Fitness landscape analyses help us to understand a problem's solution space and often predict which kinds of evolutionary search algorithms will be effective~\citep{hernandez_suite_2022}. 
Using replay experiments, we could construct \textit{potentiation} landscapes for particular evolutionary outcomes (e.g., problem-solving success) that map genotypes to evolutionary potential.
For example, mapping genotypes to problem-solving potential or potential for long-term fitness increases would create an evolvability map \citep{sansonComputationalBaselineChanges2025}.
Such potentiation mappings are an underexplored area of study (in part due to their cost to create) that we expect to yield valuable insights into how our algorithms steer populations through search spaces.

\begin{acknowledgement}
We thank the participants of Genetic Programming Theory and Practice XXIII for helpful comments and suggestions that improved this work. 
This work was supported in part through computational resources and services provided by the Institute for Cyber-Enabled Research at Michigan State University.
\end{acknowledgement}

\bibliographystyle{splncs04}
\bibliography{references,ferguson}

@article{al-tameemiMicrobialDiversificationMaintained2024,
  title = {Microbial Diversification Is Maintained in an Experimentally Evolved Synthetic Community},
  author = {{Al-Tameemi}, Zahraa and {Rodr{\'i}guez-Verdugo}, Alejandra},
  year = {2024},
  month = oct,
  journal = {mSystems},
  volume = {9},
  number = {11},
  pages = {e01053-24},
  publisher = {American Society for Microbiology},
  doi = {10.1128/msystems.01053-24},
  urldate = {2026-04-11}
}

@article{blountContingencyDeterminismEvolution2018,
  title = {Contingency and Determinism in Evolution: {{Replaying}} Life's Tape},
  shorttitle = {Contingency and Determinism in Evolution},
  author = {Blount, Zachary D. and Lenski, Richard E. and Losos, Jonathan B.},
  year = {2018},
  journal = {Science},
  volume = {362},
  number = {6415},
  publisher = {American Association for the Advancement of Science}
}

@article{blountHistoricalContingencyEvolution2008,
  title = {Historical Contingency and the Evolution of a Key Innovation in an Experimental Population of {{Escherichia}} Coli},
  author = {Blount, Zachary D. and Borland, Christina Z. and Lenski, Richard E.},
  year = {2008},
  journal = {Proceedings of the National Academy of Sciences},
  volume = {105},
  number = {23},
  pages = {7899--7906},
  publisher = {National Acad Sciences}
}

@article{covertiiiExperimentsRoleDeleterious2013,
  title = {Experiments on the Role of Deleterious Mutations as Stepping Stones in Adaptive Evolution},
  author = {Covert III, Arthur W. and Lenski, Richard E. and Wilke, Claus O. and Ofria, Charles},
  year = {2013},
  journal = {Proceedings of the National Academy of Sciences},
  volume = {110},
  number = {34},
  pages = {E3171--E3178},
  publisher = {National Acad Sciences},
  doi = {10.1073/pnas.1313424110}
}

@article{dubieDissectingSequentialEvolution2024,
  title = {Dissecting the Sequential Evolution of a Selfish Mitochondrial Genome in {{Caenorhabditis}} Elegans},
  author = {Dubie, Joseph J. and Katju, Vaishali and Bergthorsson, Ulfar},
  year = {2024},
  month = sep,
  journal = {Heredity},
  volume = {133},
  number = {3},
  pages = {186--197},
  publisher = {Nature Publishing Group},
  issn = {1365-2540},
  doi = {10.1038/s41437-024-00704-2},
  urldate = {2026-04-11},
  copyright = {2024 The Author(s)},
  langid = {english}
}

@inproceedings{fergusonPotentiatingMutationsFacilitate2023,
  title = {Potentiating {{Mutations Facilitate}} the {{Evolution}} of {{Associative Learning}} in {{Digital Organisms}}},
  booktitle = {{{Proceedings}} of the 2023 {{Artificial Life Conference}}},
  author = {Ferguson, Austin J. and Ofria, Charles},
  year = {2023},
  month = jul,
  publisher = {MIT Press},
  doi = {10.1162/isal\_a\_00684},
  urldate = {2024-03-26},
  langid = {english},
}

@inproceedings{fergusonPredictingUnpredictableUsing2024,
  title = {Predicting the {{Unpredictable}}: {{Using}} Replay Experiments to Disentangle How Evolutionary Outcomes Are Altered by Adaptive Momentum},
  shorttitle = {Predicting the {{Unpredictable}}},
  booktitle = {{{Proceedings}} of the 2024 {{Artificial Life Conference}}},
  author = {Ferguson, Austin J. and Ofria, Charles and Bohm, Clifford},
  year = {2024},
  month = jul,
  publisher = {MIT Press},
  doi = {10.1162/isal\_a\_00801},
  urldate = {2025-05-03},
  langid = {english}
}

@book{gouldWonderfulLifeBurgess1990,
  title = {Wonderful Life: The {{Burgess Shale}} and the Nature of History},
  shorttitle = {Wonderful Life},
  author = {Gould, Stephen Jay},
  year = {1990},
  publisher = {WW Norton \& Company}
}

@article{guptaHostparasiteCoevolutionPromotes2022a,
  title = {Host-Parasite Coevolution Promotes Innovation through Deformations in Fitness Landscapes},
  author = {Gupta, Animesh and Zaman, Luis and Strobel, Hannah M and Gallie, Jenna and Burmeister, Alita R and Kerr, Benjamin and Tamar, Einat S and Kishony, Roy and Meyer, Justin R},
  editor = {Ogbunugafor, C Brandon and Walczak, Aleksandra M and Barr, Jeremy},
  year = {2022},
  month = jul,
  journal = {eLife},
  volume = {11},
  pages = {e76162},
  publisher = {eLife Sciences Publications, Ltd},
  issn = {2050-084X},
  doi = {10.7554/eLife.76162},
  urldate = {2026-04-11}
}

@article{jochumsenEvolutionAntimicrobialPeptide2016a,
  title = {The Evolution of Antimicrobial Peptide Resistance in {{Pseudomonas}} Aeruginosa Is Shaped by Strong Epistatic Interactions},
  author = {Jochumsen, Nicholas and Marvig, Rasmus L. and Damki{\ae}r, S{\o}ren and Jensen, Rune Lyngklip and Paulander, Wilhelm and Molin, S{\o}ren and Jelsbak, Lars and Folkesson, Anders},
  year = {2016},
  month = oct,
  journal = {Nature Communications},
  volume = {7},
  number = {1},
  pages = {13002},
  publisher = {Nature Publishing Group},
  issn = {2041-1723},
  doi = {10.1038/ncomms13002},
  urldate = {2023-03-13},
  copyright = {2016 The Author(s)},
  langid = {english}
}

@article{lehmanSurprisingCreativityDigital2020,
  title = {The {{Surprising Creativity}} of {{Digital Evolution}}: {{A Collection}} of {{Anecdotes}} from the {{Evolutionary Computation}} and {{Artificial Life Research Communities}}},
  shorttitle = {The {{Surprising Creativity}} of {{Digital Evolution}}},
  author = {Lehman, Joel and et al.},
  year = {2020},
  month = may,
  journal = {Artificial Life},
  volume = {26},
  number = {2},
  pages = {274--306},
  issn = {1064-5462, 1530-9185},
  doi = {10.1162/artl\_a\_00319},
  urldate = {2021-08-04},
  langid = {english}
}

@article{lenskiDynamicsAdaptationDiversification1994,
  title = {Dynamics of Adaptation and Diversification: A 10,000-Generation Experiment with Bacterial Populations.},
  shorttitle = {Dynamics of Adaptation and Diversification},
  author = {Lenski, R E and Travisano, M},
  year = {1994},
  month = jul,
  journal = {Proceedings of the National Academy of Sciences},
  volume = {91},
  number = {15},
  pages = {6808--6814},
  publisher = {Proceedings of the National Academy of Sciences},
  doi = {10.1073/pnas.91.15.6808},
  urldate = {2026-05-04}
}

@article{meyerRepeatabilityContingencyEvolution2012,
  title = {Repeatability and Contingency in the Evolution of a Key Innovation in Phage Lambda},
  author = {Meyer, Justin R. and Dobias, Devin T. and Weitz, Joshua S. and Barrick, Jeffrey E. and Quick, Ryan T. and Lenski, Richard E.},
  year = {2012},
  journal = {Science},
  volume = {335},
  number = {6067},
  pages = {428--432},
  publisher = {American Association for the Advancement of Science}
}

@inproceedings{rennerComputationalModelDevelopmental2024,
  title = {A {{Computational Model}} of {{Developmental Exaptations}}},
  booktitle = {{{Proceedings}} of the 2024 {{Artificial Life Conference}}},
  author = {Renner, Alexa and Marois, {\'E}mile and Fiorito, Julian and Ashworth, Jacob and Yoder, Jason A.},
  year = {2024},
  month = jul,
  publisher = {MIT Press},
  doi = {10.1162/isal\_a\_00708},
  urldate = {2025-05-03},
  langid = {english}
}

@inproceedings{sansonComputationalBaselineChanges2025,
  title = {A Computational Baseline for Changes in Long-Term Evolvability},
  booktitle = {{{Proceedings}} of the {{Artificial Life Conference}} 2025},
  author = {Sanson, Marcos and Ferguson, Austin J.},
  year = {2025},
  month = oct,
  publisher = {MIT Press},
  doi = {10.1162/ISAL.a.895},
  urldate = {2026-02-01},
  langid = {english}
}

@article{turnerReplayingEvolutionTest2015,
  title = {Replaying Evolution to Test the Cause of Extinction of One Ecotype in an Experimentally Evolved Population},
  author = {Turner, Caroline B. and Blount, Zachary D. and Lenski, Richard E.},
  year = {2015},
  journal = {PLoS One},
  volume = {10},
  number = {11},
  pages = {e0142050},
  publisher = {Public Library of Science San Francisco, CA USA}
}

@article{vignognaExploringLocalGenetic2021,
  title = {Exploring a Local Genetic Interaction Network Using Evolutionary Replay Experiments},
  author = {Vignogna, Ryan C. and Buskirk, Sean W. and Lang, Gregory I.},
  year = {2021},
  journal = {Molecular biology and evolution},
  volume = {38},
  number = {8},
  pages = {3144--3152},
  publisher = {Oxford University Press}
}

@article{woodsSecondorderSelectionEvolvability2011,
  title = {Second-Order Selection for Evolvability in a Large {{Escherichia}} Coli Population},
  author = {Woods, Robert J. and Barrick, Jeffrey E. and Cooper, Tim F. and Shrestha, Utpala and Kauth, Mark R. and Lenski, Richard E.},
  year = {2011},
  journal = {Science},
  volume = {331},
  number = {6023},
  pages = {1433--1436},
  publisher = {American Association for the Advancement of Science}
}

@article{yedidHistoricalContingentFactors2008,
  title = {Historical and Contingent Factors Affect Re-Evolution of a Complex Feature Lost during Mass Extinction in Communities of Digital Organisms},
  author = {Yedid, Gabriel and Ofria, C. A. and Lenski, Richard E.},
  year = {2008},
  journal = {Journal of evolutionary biology},
  volume = {21},
  number = {5},
  pages = {1335--1357},
  publisher = {Wiley Online Library}
}

@inproceedings{helmuthGeneralProgramSynthesis2015,
  title = {General {{Program Synthesis Benchmark Suite}}},
  booktitle = {Proceedings of the 2015 {{Annual Conference}} on {{Genetic}} and {{Evolutionary Computation}}},
  author = {Helmuth, Thomas and Spector, Lee},
  year = {2015},
  month = jul,
  series = {{{GECCO}} '15},
  pages = {1039--1046},
  publisher = {Association for Computing Machinery},
  address = {New York, NY, USA},
  doi = {10.1145/2739480.2754769},
  urldate = {2026-05-15},
  isbn = {978-1-4503-3472-3}
}

@article{helmuthSolvingUncompromisingProblems2015,
  title = {Solving {{Uncompromising Problems With Lexicase Selection}}},
  author = {Helmuth, Thomas and Spector, Lee and Matheson, James},
  year = {2015},
  month = oct,
  journal = {IEEE Transactions on Evolutionary Computation},
  volume = {19},
  number = {5},
  pages = {630--643},
  issn = {1941-0026},
  doi = {10.1109/TEVC.2014.2362729},
  urldate = {2026-05-15}
}

@software{supplement,
  author       = {Austin Ferguson and
                  Alex Lalejini},
  title        = {Supplemental Material, GitHub repo: FergusonAJ/gp-replays},
  month        = aug,
  year         = 2026,
  publisher    = {Zenodo},
  version      = {1.0.0},
  doi          = {10.5281/zenodo.21854327},
  swhid        = {swh:1:dir:c20a532b4ac4d4419db0a58b37997d3964a7ecc1
                   ;origin=https://doi.org/10.5281/zenodo.21854326;vi
                   sit=swh:1:snp:383c873f76a21d4faed87f85718539c1681f
                   b322;anchor=swh:1:rel:e6396d332be353339de4c8bff33e
                   763066a4fed6;path=FergusonAJ-gp-replays-3044792
                  },
}

@article{lenski_convergence_2017,
	title = {Convergence and {Divergence} in a {Long}-{Term} {Experiment} with {Bacteria}},
	volume = {190},
	issn = {0003-0147, 1537-5323},
	url = {https://www.journals.uchicago.edu/doi/10.1086/691209},
	doi = {10.1086/691209},
	language = {en},
	number = {S1},
	urldate = {2026-05-14},
	journal = {The American Naturalist},
	author = {Lenski, Richard E.},
	month = aug,
	year = {2017},
	pages = {S57--S68},
}

@article{dolson_interpreting_2020,
	title = {Interpreting the {Tape} of {Life}: {Ancestry}-{Based} {Analyses} {Provide} {Insights} and {Intuition} about {Evolutionary} {Dynamics}},
	volume = {26},
	copyright = {All rights reserved},
	issn = {1064-5462, 1530-9185},
	shorttitle = {Interpreting the {Tape} of {Life}},
	url = {https://www.mitpressjournals.org/doi/abs/10.1162/artl\_a\_00313},
	doi = {10.1162/artl\_a\_00313},
	language = {en},
	number = {1},
	urldate = {2021-01-07},
	journal = {Artificial Life},
	author = {Dolson, Emily and Lalejini, Alexander and Jorgensen, Steven and Ofria, Charles},
	month = apr,
	year = {2020},
	pages = {58--79},
}

@techreport{dolson_ecological_2018,
	type = {preprint},
	title = {Ecological theory provides insights about evolutionary computation},
	url = {https://peerj.com/preprints/27315},
	doi = {10.7287/peerj.preprints.27315v1},
	language = {en},
	urldate = {2019-05-05},
	institution = {PeerJ Preprints},
	author = {Dolson, Emily L and Banzhaf, Wolfgang and Ofria, Charles},
	month = nov,
	year = {2018},
}

@incollection{ofria_avida_2009,
	address = {London},
	title = {Avida: {A} {Software} {Platform} for {Research} in {Computational} {Evolutionary} {Biology}},
	isbn = {978-1-84882-284-9 978-1-84882-285-6},
	url = {http://link.springer.com/10.1007/978-1-84882-285-6\_1},
	doi = {10.1007/978-1-84882-285-6\_1},
	language = {en},
	urldate = {2019-08-09},
	booktitle = {Artificial {Life} {Models} in {Software}},
	publisher = {Springer London},
	author = {Ofria, Charles and Bryson, David M. and Wilke, Claus O.},
	editor = {Komosinski, Maciej and Adamatzky, Andrew},
	year = {2009},
	pages = {3--35}
}

@inproceedings{lalejini_gene_2017,
	address = {Lyon, France},
	title = {Gene duplications drive the evolution of complex traits and regulation},
	copyright = {All rights reserved},
	isbn = {978-0-262-34633-7},
	url = {https://www.mitpressjournals.org/doi/abs/10.1162/isal\_a\_045},
	doi = {10.7551/ecal\_a\_045},
	language = {en},
	urldate = {2019-08-13},
	booktitle = {Proceedings of the 14th {European} {Conference} on {Artificial} {Life} {ECAL} 2017},
	publisher = {MIT Press},
	author = {Lalejini, Alexander and Wiser, Michael J. and Ofria, Charles},
	month = sep,
	year = {2017},
	pages = {257--264}
}

@article{doncieux_evolutionary_2015,
	title = {Evolutionary {Robotics}: {What}, {Why}, and {Where} to},
	volume = {2},
	issn = {2296-9144},
	shorttitle = {Evolutionary {Robotics}},
	url = {http://www.frontiersin.org/Evolutionary\_Robotics/10.3389/frobt.2015.00004},
	doi = {10.3389/frobt.2015.00004},
	urldate = {2019-08-05},
	journal = {Frontiers in Robotics and AI},
	author = {Doncieux, Stephane and Bredeche, Nicolas and Mouret, Jean-Baptiste and Eiben, Agoston E. (Gusz)},
	month = mar,
	year = {2015}
}

@article{petke_genetic_2018,
	title = {Genetic {Improvement} of {Software}: {A} {Comprehensive} {Survey}},
	volume = {22},
	copyright = {https://creativecommons.org/licenses/by/3.0/legalcode},
	issn = {1089-778X, 1089-778X, 1941-0026},
	shorttitle = {Genetic {Improvement} of {Software}},
	url = {https://ieeexplore.ieee.org/document/7911210/},
	doi = {10.1109/TEVC.2017.2693219},
	number = {3},
	urldate = {2026-08-03},
	journal = {IEEE Transactions on Evolutionary Computation},
	author = {Petke, Justyna and Haraldsson, Saemundur O. and Harman, Mark and Langdon, William B. and White, David R. and Woodward, John R.},
	month = jun,
	year = {2018},
	pages = {415--432},
}

@article{ehrgott_fifty_2026,
	title = {Fifty years of multi-objective optimization and decision-making: {From} mathematical programming to evolutionary computation},
	volume = {330},
	issn = {03772217},
	shorttitle = {Fifty years of multi-objective optimization and decision-making},
	url = {https://linkinghub.elsevier.com/retrieve/pii/S0377221725004849},
	doi = {10.1016/j.ejor.2025.06.012},
	language = {en},
	number = {1},
	urldate = {2026-08-03},
	journal = {European Journal of Operational Research},
	author = {Ehrgott, Matthias and Köksalan, Murat and Kadziński, Miłosz and Deb, Kalyanmoy},
	month = apr,
	year = {2026},
	pages = {1--25},
}

@misc{escondo_summarizing_2026,
	title = {Summarizing {Populations}: {Characterizing} the {Effects} of {Sampling} in {Computational} {Evolutionary} {Replay} {Experiments}},
	copyright = {https://creativecommons.org/licenses/by/4.0/legalcode},
	shorttitle = {Summarizing {Populations}},
	url = {https://ecoevorxiv.org/repository/view/13054/},
	doi = {10.32942/X22X0J},
	urldate = {2026-08-03},
	publisher = {Evolution},
	author = {Escondo, Nikolai and Ferguson, Austin},
	month = may,
	year = {2026},
}

@misc{hernandez_suite_2022,
    title = {A suite of diagnostic metrics for characterizing selection schemes},
    copyright = {Creative Commons Attribution Non Commercial Share Alike 4.0 International},
    url = {https://arxiv.org/abs/2204.13839},
    doi = {10.48550/ARXIV.2204.13839},
    urldate = {2024-11-06},
    publisher = {arXiv},
    author = {Hernandez, Jose Guadalupe and Lalejini, Alexander and Ofria, Charles},
    year = {2022},
    note = {Version Number: 3},
}

\end{document}